\pdfoutput=1

\documentclass[11pt]{article}

\usepackage[]{acl}

\usepackage{times}
\usepackage{latexsym}
\usepackage{multirow}
\usepackage{url}
\usepackage{amsfonts}
\usepackage{amssymb}
\usepackage{mathtools}
\usepackage{amsthm}
\usepackage{subfigure}
\usepackage{color}
\definecolor{Blue}{rgb}{0,0,1} 

\usepackage[T1]{fontenc}

\usepackage[utf8]{inputenc}

\usepackage{microtype}

\title{A Logically Consistent Chain-of-Thought Approach for Stance Detection}

\author{
Bowen Zhang\textsuperscript{1}, 
Daijun Ding\textsuperscript{1}, 
\textbf{Liwen Jing\textsuperscript{2}}
Hu Huang\textsuperscript{3}
\thanks{\ \ Emails: zhang\_bo\_wen@foxmail.com, ljing@connect.ust.hk, h.huang@pku.edu.cn}\\
\textsuperscript{1} College of Big Data and Internet, Shenzhen Technology University, Shenzhen, China\\
\textsuperscript{2} Shenzhen X-Institute, Shenzhen, China \\
\textsuperscript{3} Peking University Shenzhen Graduate School, Shenzhen, China \\
  }

\begin{document}
\maketitle
\begin{abstract}
Zero-shot stance detection (ZSSD) aims to detect stances toward unseen targets. 
Incorporating background knowledge to enhance transferability between seen and unseen targets constitutes the primary approach of ZSSD.
However, these methods often struggle with a knowledge-task disconnect and lack logical consistency in their predictions.  
To address these issues, we introduce a novel approach named Logically Consistent Chain-of-Thought (LC-CoT) for ZSSD, which improves stance detection by ensuring relevant and logically sound knowledge extraction.
LC-CoT employs a three-step process. Initially, it assesses whether supplementary external knowledge is necessary. Subsequently, it uses API calls to retrieve this knowledge, which can be processed by a separate LLM. Finally, a manual exemplar guides the LLM to infer stance categories, using an \textit{if-then} logical structure to maintain relevance and logical coherence. This structured approach to eliciting background knowledge enhances the model's capability, outperforming traditional supervised methods without relying on labeled data. 
\end{abstract}

\section{Introduction}

Stance detection is a fundamental natural language processing (NLP) task that categorizes expressed attitudes toward a particular target based on opinionated input texts \cite{kuccuk2020stance,ding2024cross}. This task has attracted significant research attention in recent years due to its relevance across domains like political analysis, social media monitoring, and customer feedback analysis \cite{yang2019investigating, ZhangTSMWLP21}.
In practice, the enumeration of all conceivable targets in advance for training stance detection models is infeasible.
Consequently, zero-shot stance detection (ZSSD) has emerged as a promising approach, focused on accurately identifying the stance towards unseen targets during the inference stage \cite{allaway2020zero}.

ZSSD is traditionally framed as a target-based sentence-level classification task that utilizes either non-pretrained or pretrained language models (PLMs) \citet{jiang2019hierarchical,liang2022zero}. However, sentences often contain background knowledge such as domain-specific terminology, cultural references, social media linguistic styles, and more. These elements are not readily comprehensible to conventional methods and require specialized parsing to be fully understood.

Recently, efforts to improve ZSSD have focused on the exploitation of such background knowledge, predominantly through unsupervised methods owing to the scarcity of explicitly annotated background data \citet{liu2021enhancing,zhu2022enhancing,liang2022jointcl}. 
The emergence of large language models (LLMs), such as GPT series, trained on comprehensive text corpora, presents new avenues for knowledge extraction to bolster stance detection \cite{li2023stance, zhang2023investigating}.
However, current ZSSD approaches exhibit clear deficiencies in knowledge utilization, leading to two key issues:
1) Knowledge-task disconnect: conventional approaches tend to extract expansive, fragmented information components that have limited relevance to the specific stance detection task. This can impair performance when processing contextual data that is highly interdependent with the target stance.
2) Lack of logical consistency: the fragmented knowledge lacks necessary logical verification, introducing potential errors and contradictions that diminish the credibility of stance predictions.

To achieve this goal, in this paper, we propose a \textbf{L}ogically \textbf{C}onsistent \textbf{C}hain-of-\textbf{T}hought (LC-CoT) Approach for ZSSD.
LC-CoT approaches involve utilizing manually-designed prompt templates to extract background knowledge relevant to the stance detection analysis process from LLMs in a CoT manner.
Specifically, LC-CoT consists of three steps.
First, we ask the LLM to determine if additional external knowledge is required for the given input. 
Second, the LLM is leveraged to produce knowledge retrieval APIs through utilizing tools via API invocations. This API can feed into a separate LLM to obtain knowledge. Third, we furnish the LLM with a manual selected exemplar to guide the LLM in inferring stance categories by consolidating the input and background knowledge. In this step, the generated template follows \textit{if-then} logical structures~\cite{zhang2022sentiment} to ensure knowledge utilization and inference aligns with the stance prediction process.
We conducted extensive experiments validating that eliciting background knowledge following \textit{if-then} format can effectively augment model capabilities to surpass traditional supervised approaches even without labeled samples. 

\section{Method}


\textbf{Task Definition and Model Overview.}
We use $D=\{x_{i}, p_{i}\}$ to denote the collection of input data, where $x$ and $p$ denote the input text and the corresponding target, respectively. 
The stance detection task aims to predict a stance label $\hat{y}_i$ for given input $\{x_{i}, p_{i}\}$. 

\subsection{LC-CoT} 

To address the challenge of knowledge-task disconnect and lack logical consistency, and leverage the rich knowledge encoded within LLMs, we designed a three step CoT method. In the first step, we engage an LLM to ascertain the necessity for additional external knowledge pertinent to the input text. Upon establishing this need, in the second step, we exploit the LLM's capabilities to generate API calls, effectively creating a bridge to external knowledge bases. This facilitates the acquisition of pertinent information from a distinct LLM, tailored to the context of the stance detection task. 
In the third step, the LLM is provided with a carefully chosen exemplar, which serves as a cognitive scaffold, directing the inference process. By employing \textit{if-then} logical constructs within the generated templates, we ensure that the assimilation of input with the procured background knowledge is both relevant and logically consistent, thereby enhancing the accuracy and reliability of the stance categorization.

\textbf{Step1}: We first feed the constructed instruction $S'1$ into the LLM to decide whether external knowledge is required for stance prediction.
\begin{center}
\fcolorbox{black}{blue!10}{\parbox{0.95\linewidth}{$S'1$: \textit{Your task is to judge whether there is enough evidence to support the stance prediction based on the text content.}\\
\textbf{Input}:  [input text $x$] to the target [given target $p$].\\
\textbf{Output}: [yes/no]}}
\end{center}

\textbf{Step2}: 
If the model output requires additional information, the $S'2$ instruction can be used to allow the LLM to automatically retrieve the required background knowledge ($q$):
\begin{center}
\fcolorbox{black}{blue!10}{\parbox{0.95\linewidth}{$S'2$:
\textit{You can call the API by writing "QUERY [A]" where "A" is the required knowledge. 
Here are some examples of API calls:}\\
\textbf{Input}: What's the attitude of the sentence [input text $x$] to the target [given target $o$]? Select an answer from (favor, against, none) or API call.\\
\textbf{Output}: API call, QUERY [$\dots$]}}
\end{center}

\textbf{Step 3}: 
Finally, we send the API call $S'3$ to the LLM to acquire the \textit{if-then} expression.

\begin{center}
\fcolorbox{black}{blue!10}{\parbox{0.95\linewidth}{$S'3$:
\textit{Your task is to add calls to a Question Answering API to a piece of text. The questions should help you get information required to complete the text. You can call the API by writing "[RULE: IF (A) then (B)]" where "A" is the reason why "B". 
Here are some examples of API calls: }\\
\textbf{Input}: what's the attitude of the sentence [input text $x$] to the target [given target $o$]? [given knowledge $q$ (if have)]. select an answer from (favor, against, none). \\
\textbf{Output}: [IF (reason) then (attitude is [stance label])]. \\
Input: $\dots$
}
}
\end{center}

\begin{figure*}[htbp]
	\centering
	\includegraphics[width=\linewidth]{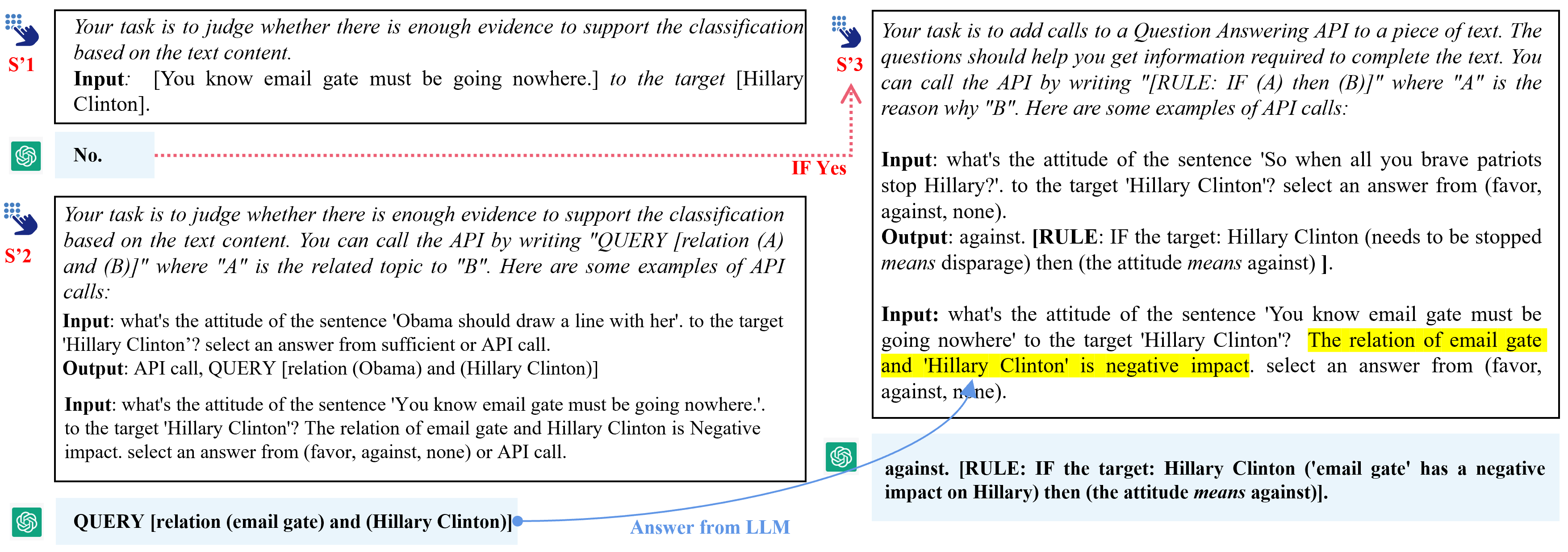}
	\caption{An illustration of LC-CoT.}
 \label{example11}
\end{figure*}

Figure \ref{example11} shows an example, given the input as: ``\textit{You know email gate must be going nowhere.}'' to the target ``\textit{Hillary Clinton}'', the LC-CoT model can generate a output: ``\textit{IF the target: Hillary Clinton (`email gate' has a negative impact on Hillary) then (the attitude is against)}''.
Ultimately, we can extract \textit{against} from the \textit{if-then} expression, that can serve as the stance prediction label for the LC-CoT model.

\section{Experiment}

\subsection{Experimental Data}
This paper presents experimental results on robust benchmark datasets, including SemEval-2016 Task 6 (SEM16) \cite{StanceSemEval2016} and VAST \cite{allaway2020zero}.
{SEM16} comprises 4870 tweets with diverse targets, with each tweet being labeled with a label of ``\textit{favor}'', ``\textit{against}’’, or ``\textit{neutral}’’. In accordance with the proposed configuration by \cite{wei2019modeling}, four targets, i.e., \textit{Donald Trump} (D), \textit{Hillary Clinton} (H), \textit{Legalization of Abortion} (L), and \textit{Feminist Movement} (F), are selected for evaluating the efficacy of the stance detection task, and hence have been chosen for our study.
Following \cite{allaway2021adversarial}, we regard a target as the zero-shot testing target while training on the other five, and randomly select 15\% of the training set as the development data to tune the hyper-parameters.
VAST comprises three distinct stance labels, with the label set defined as {``Pro'', ``Neutral'', ``Con''}. The training set comprises 4003 samples, while the dev and test sets consist of 383 and 600 samples, respectively.

\subsection{Compared Baseline Methods}
To assess the efficacy of our proposed model, we conduct a thorough evaluation and comparison with a range of established baselines. The details of these baseline models are presented below for reference:

\textbf{Statistics-based methods.} {BiCond} \cite{augenstein2016stance} utilized a bidirectional-LSTM to encode the underlying sentence and the corresponding target. 
{CrossNet} \cite{xu2018cross} is a variant of BiCond, which leverages a self-attention layer to capture informative words.
{TPDG} \cite{LiangF00DHX21} proposed a target-adaptive graph convolutional network. 
{AT-JSS-Lex} \cite{li2019multi} developed a target-adaptive graph convolutional network for the purpose of stance detection. 
{SEKT} \cite{bowenacl} introduced semantic knowledge as the transferable knowledge between targets.

\textbf{Fine-tuning based methods.} {Bert-FT} \cite{Bert} employed a pretrained BERT model for stance detection.  
{PT-HCL} \cite{liang2022zero} developed a novel approach to cross-target and zero-shot stance detection using contrastive learning.
JointCL \cite{liang2022jointcl} proposed a contrastive learning method to leverage the stance features of known targets.
TarBK \cite{zhu2022enhancing} incorporated the targeted background knowledge for stance detection.
TTS \cite{li2023tts} proposed to augment the training set with different diverse targets.

\textbf{LLM-based methods.} 
GPT-DQA \cite{zhang2022would} directly elicited stance categories from GPT-3.5 by posing queries in an interrogative format.
GPT-CoT \cite{zhang2023investigating} developed the method that prompts LLM with artificially constructed examples containing predefined inferential logic in a chain-of-thought manner to obtain stance categories. 
KASD-GPT \cite{li2023stance} utilized GPT-3.5 to retrieve relevant background knowledge, which is then integrated into the trainable Bert to exploit annotated samples through backpropagation.


\begin{table}[htbp]
\small
\begin{center}
\begin{tabular}{l|cccc|c}
\hline
\multirow{2}{*}{Model}    & \multicolumn{4}{c|}{SEM16} & \multicolumn{1}{c}{VAST}  \\ \cline{2-6}
  & HC   & FM   & LA & DT & All \\ \hline
Bicond   & 32.7 & 40.6 & 34.4 & 30.5  & 41.0  \\
CrossNet  & 38.3 & 41.7 & 38.5 &35.6  & 45.5  \\
SEKT     & 50.1 & 44.2 & 44.6 & 46.8  &  41.1  \\
TPDG   & 50.9 & 53.6 & 46.5 & 47.3 &  51.9 \\
\hline
Bert-FT      & 49.6 & 41.9 & 44.8 & 40.1 &  65.3  \\
PT-HCL   & 54.5 & 54.6 & 50.9 & 50.1  & 71.6 \\ 
JointCL & 54.8 & 53.8 & 49.5 & 50.5  & 72.3 \\
TarBK &  55.1 & 53.8 & 48.7 & -   & \textbf{73.6}  \\
\hline 
GPT-DQA  &   {78.0}    &  {69.0}  &   {59.3} & {71.3}   &62.3 \\
GPT-CoT  &   {78.9}    &  68.7  &   {61.5} & {71.6}  & 68.9 \\
KASD-ChatGPT & 80.3 & \textbf{70.4} & 62.7 &-&
67.0\\
\hline 
LC-CoT (Ours) & \textbf{82.9} &  \textbf{70.4}   &   \textbf{63.2}  &  \textbf{71.7}   & 	72.5  \\ 
\hline 	
\end{tabular}
\end{center}
\caption{Zero-shot stance detection experiment results. The best scores are in bold.
}
\label{tab1}
\end{table}

\begin{table}[!h]
\begin{center}	
\small
\begin{tabular}{l|l|cccc}
\hline 
&Methods & F$\to$L  & L$\to$F & H$\to$D & D$\to$H  \\ 
\hline 
\multirow{6}{*}{Sta.} & BiLSTM $\dag$         & 44.8  & 41.2 & 29.8 & 35.8   \\
&BiCond        & 45.0   & 41.6 & 29.7 & 35.8  \\
&CrossNet      & 45.4  & 43.3 & 43.1   & 36.2   \\
&VTN           & 47.3  & 47.8 & 47.9 & 36.4  \\
&SEKT          & 53.6 & 51.3 &  47.7   &{42.0}    \\
&TPDG          & {58.3}  & 54.1 & 50.4 & 52.9   \\ 
\hline
\multirow{6}{*}{BERT} & BERT-FT       & 47.9  & 33.9 & 43.6 & 36.5 \\      
&JointCL & 58.8  & 54.5& 52.8 &54.3 \\
&PT-HCL & 59.3 & 54.6 & 53.7 & 55.3 \\
&TarBK &  59.1&  54.6&   53.1&  54.2 \\
\hline
& LC-CoT (Ours) &  \textbf{63.2}  &  \textbf{70.4} & \textbf{71.7}  &  \textbf{82.9}  \\  
\hline		
	\end{tabular}
\caption{Cross-target stance detection experiment results.}
\end{center}
\label{crosst}
\end{table}


\section{Experiment}
We report the main experimental results of zero-shot stance detection in Table \ref{tab1}. 
Following previous methods, we adopted macro-averaged F1 ($F1_{m} =(F1_{micro}+F1_{macro})/2$) as the evaluation metric for our study to verify our model.

We observe that our LC-CoT performs consistently better than most of the baseline models on all datasets, which verifies the effectiveness of our proposed approach in ZSSD. 
Despite the challenging nature of ZSSD, our LC-CoT model exhibits considerable potential, surpassing all benchmark approaches on the SEM16 and VAST datasets. 
Specifically, Our LC-CoT substantially surpassing TarBK on three targets on average by 19.5\% with SEM16.
Notably, LC-CoT exhibits a slightly inferior performance compared to TarBK,  the best contrastive model incorporating knowledge conventionally.
However, LC-CoT does not necessitate any training data, unlike TarBK which necessitates substantial labeled data for training, amply demonstrating its superior capability in acquiring background knowledge requisite for stance detection.


Additionally, our LC-CoT continued to achieve valid improvements compared to CoT-based methods. Contrasted with CoT-based methods (GPT-DQA and GPT-CoT), LC-CoT exhibited significant enhancements on SEM16 and VAST datasets. 
Notably, relative to the knowledge generation model KASD-ChatGPT predicated on LLM, LC-CoT averaged improvements of 1.03\% across 3 targets on the SEM16 dataset and 5.5\% on VAST. 
Such improvements highlight the effectiveness of integrating logical reasoning into the knowledge generation process, which significantly elevates the performance of the model. The experimental results demonstrate that by aligning the CoT-based knowledge elicitation with the model's stance prediction logic, LC-CoT not only enhances accuracy but also sets a precedent for future stance detection methodologies.

\subsection{Cross-Target Setup}
To evaluate the generalizability of our LC-CoT method for cross-target stance detection, we also assessed LC-CoT under cross-target conditions on the SEM16 datasets. The objective of the cross-target configuration is to predict the stance towards the target destination utilizing labeled data from the source target.

The results are presented in Table 2. Based on these results, LC-CoT substantially outperforms the other baselines. Specifically, relative to the previously top statistical method (TPDG), LC-CoT achieves an average improvement of 18.1\% on the F1 score on average, affirming the efficacy of employing a distantly supervised framework in the cross-target setting. Compared to the best fine-tuning-based approach (PT-HCL), LC-CoT exhibits an average 16.3\% enhancement on the F1 score on average.
As GPT-DQA, GPT-CoT, and KASD-ChatGPT require no training data, their results remain consistent with Table 1 and are thus not reiterated here.

\section{Conclusion}
This paper proposes the Logically Consistent Chain-of-Thought (LC-CoT) approach, which refines knowledge extraction and application by using LLMs in a structured, logical manner.
LC-CoT operates in three stages to assure relevance and logical soundness in stance detection. First, it evaluates the need for external knowledge. Then, it retrieves this information via APIs, harnessing the comprehensive understanding capabilities of LLMs. Finally, LC-CoT employs \textit{if-then} reasoning patterns, guided by manual exemplars, to integrate the knowledge into the stance detection process. This method has proven superior to traditional supervised techniques, even in the absence of labeled training data, and when fused with existing models, significantly enhances their accuracy.
Our research validates LC-CoT as a powerful tool for ZSSD, marking a step forward in the use of LLMs for tasks requiring not just data processing but also nuanced comprehension and logical inference.

\bibliography{anthology,custom}
\bibliographystyle{acl_natbib}


\end{document}